\pdfoutput=1

\documentclass[conference]{IEEEtran}
\ifCLASSINFOpdf
\else
\fi
\usepackage{amsmath}

\usepackage{caption} 
\usepackage{booktabs}
\usepackage{enumitem} 
\usepackage{graphicx}
\usepackage{algorithm}
\usepackage{algcompatible}
\usepackage{algpseudocode}
\usepackage{amsfonts}

\usepackage[T1]{fontenc}
\usepackage[font=small,labelfont=bf,tableposition=top]{caption}

\usepackage{capt-of}

\usepackage{accents}

\usepackage{algcompatible}
\usepackage{algpseudocode}
\usepackage{pifont}

\PassOptionsToPackage{hyphens}{url}\usepackage{hyperref}

\usepackage{paracol}
\usepackage{xparse}
\usepackage{kantlipsum}

\NewDocumentCommand\myframedtext{ s O{.9\linewidth} m }{%
    \IfBooleanTF{#1}{\begin{figure*}}{\begin{figure}}%
      \centering%
      \fbox{\parbox{#2}{%
        #3%
      }}%
    \IfBooleanTF{#1}{\end{figure*}}{\end{figure}}}

\newcommand{\ds}{\displaystyle}
\newcommand{\df}{\displaystyle\frac}

\newcommand{\+}[1]{\ensuremath{{\boldsymbol #1}}} 
\newcommand{\quotes}[1]{``#1''}
\newcommand{\R}{\mathbb{R}}
\newcommand{\red}[1]{\color{red}{#1}\color{black}}

\newcommand{\blue}[1]{\color{blue}{#1}\color{black}}

\newcommand\numberthis{\addtocounter{equation}{1}\tag{\theequation}}

\makeatletter
\newcommand*\bigcdot{\mathpalette\bigcdot@{.5}}
\newcommand*\bigcdot@[2]{\mathbin{\vcenter{\hbox{\scalebox{#2}{$\m@th#1\bullet$}}}}}
\makeatother

\begin{document}
%
\title{"Influence Sketching": Finding Influential Samples In Large-Scale Regressions}


\author{\IEEEauthorblockN{Mike Wojnowicz, Ben Cruz, Xuan Zhao, Brian Wallace, Matt Wolff, Jay Luan, Caleb Crable}
\IEEEauthorblockA{Department of Research and Intelligence \\
Cylance, Inc. \\
Irvine, California 92612 \\
\{mwojnowicz, bcruz, xzhao, bwallace, mwolff, jluan, ccrable\}@cylance.com}

}

\maketitle


\begin{abstract}

There is an especially strong need in modern large-scale data analysis to prioritize samples for manual inspection.  For example, the inspection could target important mislabeled samples or key vulnerabilities exploitable by an adversarial attack.   In order to solve the ``needle in the haystack" problem of which samples to inspect, we develop a new scalable version of Cook's distance, a classical statistical technique for identifying samples which unusually strongly impact the fit of a regression model (and its downstream predictions).  In order to scale this technique up to very large and high-dimensional datasets, we introduce a new algorithm which we call ``influence sketching."   Influence sketching embeds random projections within the influence computation; in particular, the influence score is calculated using the randomly projected pseudo-dataset from the post-convergence Generalized Linear Model (GLM).   We validate that influence sketching can reliably and successfully discover influential samples by applying the technique to a malware detection dataset of over 2 million executable files, each represented with almost 100,000 features.  For example, we find that randomly deleting approximately 10\% of training samples reduces predictive accuracy only slightly from 99.47\% to 99.45\%, whereas deleting the same number of samples with high influence sketch scores reduces predictive accuracy all the way down to 90.24\%.  Moreover, we find that influential samples are especially likely to be mislabeled.  In the case study, we manually inspect the most influential samples, and find that influence sketching pointed us to new, previously unidentified pieces of malware.\footnote{This work is copyrighted by the IEEE. Personal use of this material is permitted. However, permission to reprint/republish this material for advertising or promotional purposes or for creating new collective works for resale or redistribution to servers or lists, or to reuse any copyrighted component of this work in other works must be obtained from the IEEE.} 

\end{abstract}


\IEEEpeerreviewmaketitle

\section{Introduction}


Sample influence scores have been largely neglected in modern large-scale data analysis, perhaps considered a mere anachronism in the historical context of science experiments which explored model behavior with and without a handful of aberrant cases.   However, sample influence scores can help solve the ``needle in the haystack" problem in modern data analysis: where in a corpus of many millions of samples should one devote one's attention?    Modern datasets can be of such enormous sizes that human experts may have had a small-to-nonexistent role in their creation.  However, human experts may be available to provide a deeper analysis of some but not all samples.  A scalable measure of sample influence could solve this \emph{queueing problem}, prioritizing samples which have the largest model impact.   

Consider, for example, two special cases. First, modern large-scale datasets are often labeled by a heuristic or algorithm, introducing the problem of \emph{label noise}~\cite{frenay}.   For example, in the field of cybersecurity, there are not enough professional reverse engineers to analyze every new piece of software that encounters the world's computers; thus, the security of software is traditionally determined based on whether ``signatures" of maliciousness have been satisfied.  Similarly, in the field of natural language processing, inexpensive labeling from non-experts can be often obtained for huge datasets through Mechanical Turk\footnote{\url{www.mturk.com}}; however, this labeling can be substantially less reliable than labeling by experts~\cite{frenay}.  A scalable measure of sample influence could point towards potentially mislabeled samples, particularly those with pernicious model impact.   Second, even high-performing models can harbor vulnerabilities, as in the now well-known example of the model that learned to discriminate between wolves and huskies through background ``snow features"~\cite{ribeiro}.  Because the model's decision making was based not on physical features, but on background snow, the model was vulnerable to attack.    A scalable measure of sample influence could point towards important vulnerabilities in the model, therefore subserving a primary goal of \emph{adversarial learning}.

In this paper, we present an attempt to revive, modernize, and scale up a technique from classical statistics of the late 1970s: a measure of sample influence known as Cook's Distance~\cite{cook}.  In particular, we focus on Generalized Cook's Distance~\cite{pregibon}, which can identify influential samples with respect to any regression model  (linear, logistic, Poisson, beta, etc.) in the family of Generalized Linear Models.   Cook's Distance deems a sample influential when its inclusion causes ``strange" (or unexpected) perturbations in the regression weights.  As we will see, these influential samples can have a strong impact on a model's predictions about future samples.   High influence scores are associated with samples that are hard to predict and tend to lie in unusual locations of feature space (in a way that depends upon principal component subspaces, as described more precisely below).   

 Cook's Distance has a key feature that makes it an excellent candidate for scaling up the large datasets: it can compare the fit of a regression model to a set of $n$ regression models, each of which omits one sample from the training set, without the user actually having to run $n$ separate regression models.    Indeed, in modern data analysis, running $n$ regressions can be quite difficult: there may be very many high-dimensional samples, and the chosen regression model (e.g. a logistic lasso regression) may be computationally expensive to fit.  At the same time, however, the construction of Generalized Cook's Distance involves matrix operations (especially, forming and inverting a $n \times n$ covariance matrix) that can easily render it computationally infeasible on large data sets. 
 
 To solve this problem, we present an {\bf influence sketching} procedure, which extends Generalized Cook's Distance to large scale regressions by embedding random projections~\cite{achlioptas} within its construction.   For a dataset with $n$ samples and $p$ predictors, the algorithm has approximate worst-case complexity of $\mathcal{O}(np\log(n))$, with potentially lower complexity when the dataset and/or projection matrix is sparse.  
We argue analytically, and demonstrate empirically, that ``influence sketching" can successfully identify impactful samples in large-scale regressions.    In the presence of label noise, the technique is particularly valuable; if an expert can assess ground truth with high accuracy relative to the labeling mechanism, then inspected miscategorized influential samples can logically be determined to be either (a) mislabeled samples which significantly impact the model or (b) key model deficiencies (which could spur feature development, possibly protecting against adversarial attack).   Moreover, we find that, at least on our dataset, influential samples are especially likely to be mislabeled.  By investigating highly influential samples which were nominally miscategorized by a malware classifier, we discover previously unidentified malware samples.  

\section{Preliminaries}
\subsection{Notation}
For the dataset, we assume that $\+X \in \R^{n \times p}$ is a dataset of $n$ samples and $p$ predictors, and that  $\+y \in \R^{n}$ is a vector of outcomes.

For multivariate linear regression, we assume that $\+y= \+X\+\beta +\epsilon$, where $\+\epsilon \sim N_n(\+0,\sigma^2 \+I)$.  The fitted values of the response variable are given by $\widehat{\+y} = \+X\widehat{\+\beta}$, where the ordinary least squares solution for the parameter is $\widehat{\+\beta} = (\+X^T\+X)^{-1}\+X^T\+y$.  The vector of residuals is given by $\widehat{\+r} = \+y - \widehat{\+y}$.   (More generally, in generalized linear regression, the vector of residuals is given by  $\widehat{\+r} = \+y - \widehat{\mathbb{E}}[\+y|\+X]$, where $\widehat{\mathbb{E}}[\+y|\+X]$ is the expected value of the response variable, according to the model, given the predictors.) The value of the regression weights with the $i$th sample deleted from the dataset is denoted by $\widehat{\+\beta}_{(i)}$.  Whether $\+x_i$  represents the $i$th row or $i$th column of $\+X$ should be clear from context; without a transpose symbol, $\+x_i$ is oriented as a matrix with a single column. The hat notation refers to a population parameter that is estimated from a dataset.    

For random projections, we let $\+\Omega \in \R^{p \times k}$ be a random projection matrix, where $k$ is the target or reduced dimensionality for the feature space.   We let $\+Y \in \R^{n \times k} : \+Y = \+X \+\Omega$ be the randomly projected dataset. 

\subsection{Types of Errors} \label{errors}

In this section, we introduce nomenclature to differentiate possible errors in the presence of label noise.   When labels are uncertain, errors can occur either during the \emph{modeling process} (which relates labels, $\+y$, to features, $\+X$) or during the \emph{labeling process} (which assigns labels, $\+y$, to samples). 

We refer to errors in the modeling process (i.e., where the model's prediction does not match the nominal label) as \emph{modeling errors} or \emph{nominal errors}.  In particular, a nominal Type 1 error (or nominal false positive) is the event $\{\text{model predicts bad and provided label is good} \}$, and nominal model Type 2 error (nominal miss) is the event $\{\text{model predicts good and provided label is bad} \}$.   Note that nominal modeling miscategorizations can happen at the level of training or testing.

We refer to errors in the labeling mechanism (i.e., where the nominal label does not match the actual ground truth) as \emph{labeling errors}.  In particular, a labeling Type 1 error (labeling false positive) is the event $\{\text{sample labeled as bad  and sample is actually good} \}$, and labeling Type II error (labeling miss) is the event $\{\text{sample is labeled as good and sample is actually bad} \}$.   
 
The model produces an \emph{actual miscategorization} (i.e., the model's prediction does not match the actual label) when either a modeling error or a labeling error occurs -- but not both.  The distinctions between these types of errors will be particularly important in Section~\ref{results2}.  




\section{Background}
\subsection{Leverage} \label{leverage_primer}
\subsubsection{Construction} \label{leverage_construction} 
  
The concept of influence depends, in part, on the concept of \emph{leverage}~\cite{cook_and_weisberg}.   Leverage is a particularly useful quantity for describing unusual samples, reflecting the degree to which a  sample lies in extreme locations of feature space. 

Leverage scores are computed from the hat matrix, given by
  \begin{equation}
 \label{hat_def}
 \+H=\+X(\+X^T\+X)^{-1}\+X^T
 \end{equation}
 The hat matrix is an orthogonal projection matrix which projects vectors onto the column space of $\+X$. For example, in linear regression, the hat matrix $\+H$ maps the observed values $\+y$ onto the fitted values $\widehat{\+y}$ (and so it ``puts the hat on" $\+y$).  
     
The leverage for the $i$th sample is given by $h_i$, the $i$th diagonal of the hat matrix: 
\begin{equation}
\label{hat_diagonal}
h_i = \+x_i^T (\+X^T\+X)^{-1} \+x_i
\end{equation}

Through the idempotency and symmetry of the hat matrix, $\+H$, it is possible to derive the bound $0 \leq h_i \leq 1$.

\subsubsection{Interpretation}    
\begin{figure*}
    \centering
    \includegraphics[height=2.5in]{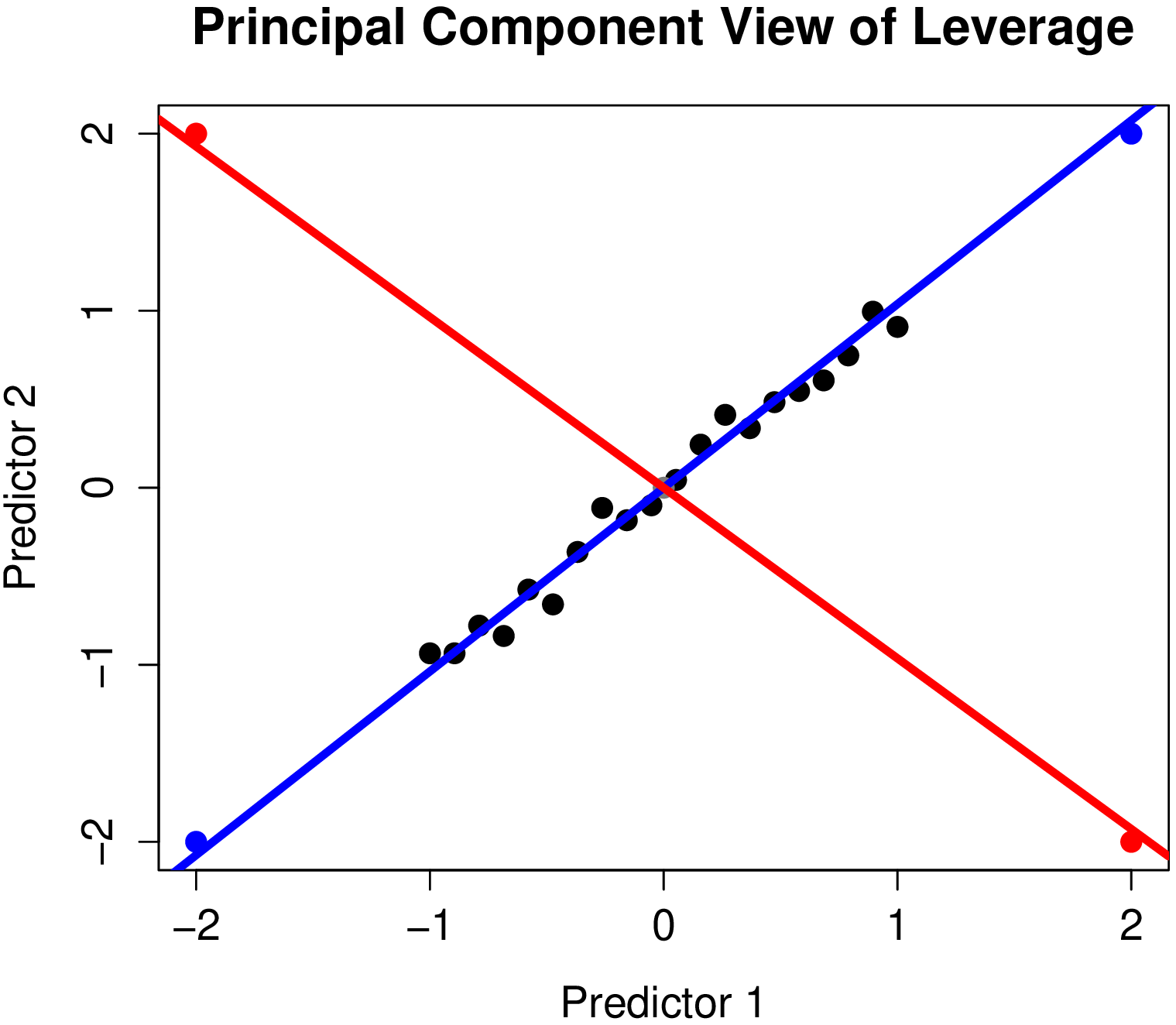}
    \qquad
    \begin{tabular}[b]{ccccc}
    \multicolumn{5}{c}{{\bf Leverage Values For Dataset}} \\
0.06&  0.04 &0.04 &0.04 &0.02 \\
0.02 &0.01 &0.00& 0.00 &0.00 \\
0.00 &0.00 &0.01& 0.01& 0.02 \\
0.02 &0.03 &0.04 &0.04& 0.07 \\
0.00 &\blue{0.27} &\blue{0.27} & \red{0.50}& \red{0.50}\\
    \end{tabular}
    \caption{\emph{A principal components view of leverage}.  The first and second principal components are plotted as lines colored blue and red, respectively.    The red and blue points lie about equally far from the mean of the dataset (about (0,0)), and they all have unusually large leverage.    However, the red data points have much larger leverage than the blue data points.  This is because they primarily lie along a non-dominant principal component.}
      \label{hi}
  \end{figure*}

We can explain how leverage quantifies whether a sample lies in ``extreme locations of feature space" by relating the statistic to the dataset's principal components.    Let $\+X^c$ represent the centered dataset and $\+x_i^c$ represent the $i$th sample in a centered dataset (i.e. the mean is subtracted off of all features).
Let the dataset $\+X^c$ be full rank, and let $\+x_{i,1}^{pca},\hdots,\+x_{i,P}^{pca}$ represent the $P$ principal components for $\+x_i^c$, listed in order of non-decreasing eigenvalues  $\lambda_1^2 \geq ... \geq \lambda_P^2$.  Using the singular value decomposition for $\+X^c$ and the fact that the principal directions are the right singular vectors, it is easy to show that:
\[ h_i = \ds\sum_{p=1}^P \bigg(\df{\+x_{i,p}^{pca}}{\lambda_p}\bigg)^2 \] 
In fact, letting $\theta_{ip}$ represent the angle between the $i$th sample and $p$th principal component in feature space $\R^p$, we obtain:
\begin{equation} 
\label{leverage_rep}
h_i = || \+x_i^c ||_2^2 \; \ds\sum_{p=1}^P \bigg(\df{cos (\theta_{ip})}{\lambda_p}\bigg)^2
\end{equation} 
Equation~\ref{leverage_rep} nicely reveals that leverage values are large if (1) the sample is far from the bulk of the other samples in feature space, and (2) a substantial part of its magnitude lies along non-dominant principal components.   For a visual representation of this analysis, see Figure~\ref{hi}.  

\subsubsection{Alternative Viewpoint} \label{leverage_alternative}
\begin{figure}
\centering
\includegraphics[height=3.2in]{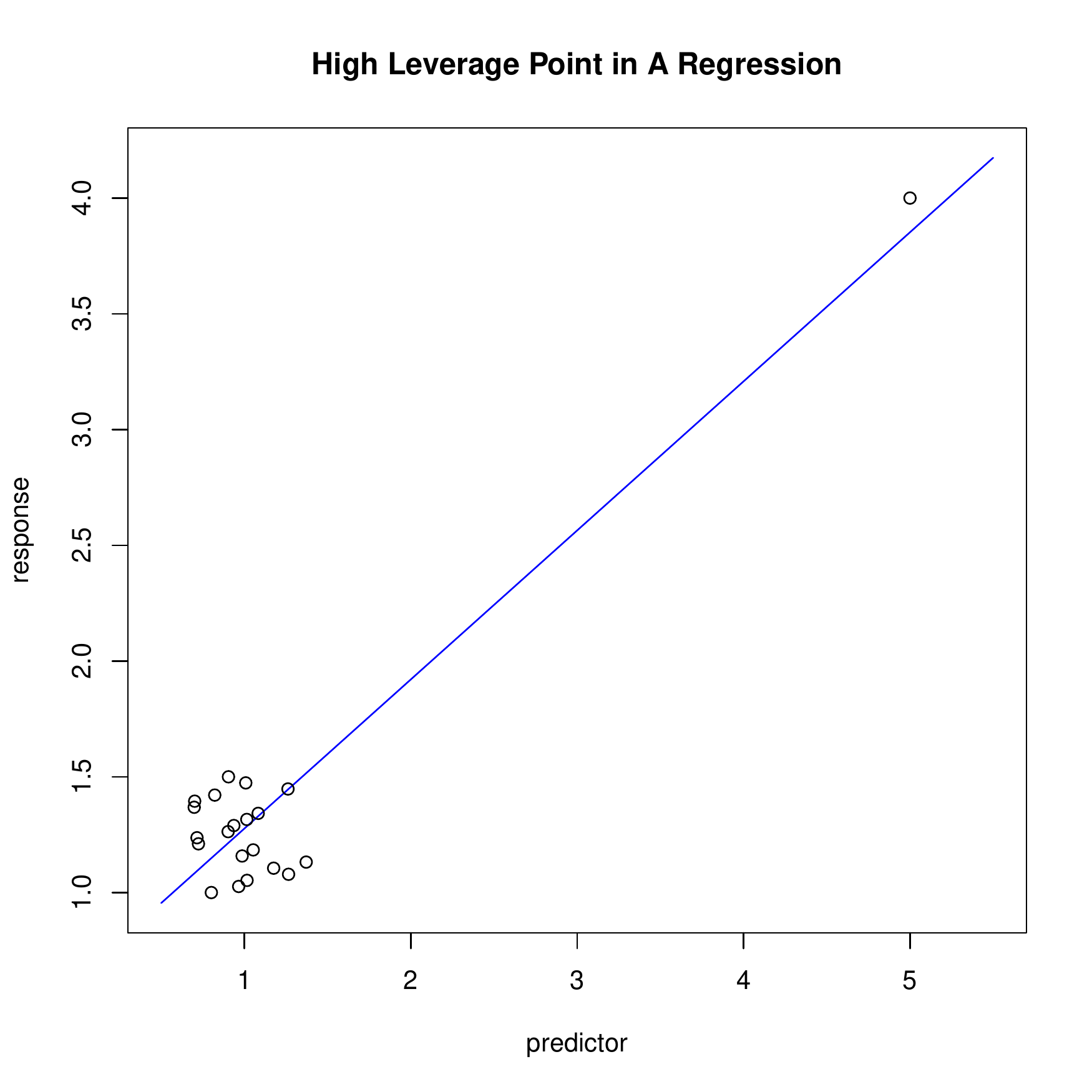}
\caption{ \emph{A high leverage point in the context of simple linear regression.} The point on the upper right of the plot has very high leverage.  If the point were pushed downward, the entire regression line would shift dramatically. }
\label{leverage_line}
\end{figure}

For a related viewpoint on leverage, consider an alternative construction: $h_i/(1-h_i)$.  Although in Section~\ref{leverage_construction}, the term leverage referred simply to the numerator, $h_i$, the two terms are monotonically related. Thus, both formulations of leverage provide an identical ordering of samples.  To interpret this variant, consider that using the idempotency of $\+H$ as an orthogonal projection, we can obtain formulas for the variance of the fitted values and residuals in the case of linear regression:   
\begin{equation}
\label{var_fitted}
Var(\widehat{\+y})=  \+X (\+X^T \+X)^{-1} \+X^T \sigma^2 = \+H \sigma^2 
\end{equation}
\begin{equation}
\label{var_residuals}
Var(\widehat{\+r})= \bigg(\+I - \+X (\+X^T \+X)^{-1} \+X^T\bigg) \sigma^2 = (\+I - \+H) \sigma^2   
\end{equation}
Based on these formulas, we obtain
\[   \df{ h_i}{(1-h_i)}  = \df{Var(\widehat{y}_i)}{Var(\widehat{r}_i)}\]
For a simple illustration, Figure~\ref{leverage_line} captures how a simple linear regression model tends to provide excellent fits to extreme points in predictor space, tracking them as they move around.   As a result, these extreme points have small variance for the residuals, $Var(\widehat{r}_i)$, relative to the variance for the fitted values, $Var(\widehat{y}_i)$.  Thus, outlying samples are called high ``leverage" because the fitted regression model is particularly sensitive to them.
 
\subsection{Cook's Distance: A measure of influence} \label{influence}
\subsubsection{Construction} \label{cooks_stat}
Cook's distance, $c_i$, was developed for regression models in order to summarize the influence of the $i$th sample on the model fit~\cite{cook}.   Cook's distance accomplishes this through \emph{case deletion}; i.e., by comparing the fitted regression weights if all samples are fit, versus if all but the $i$th sample are fit.   This comparison is $(\widehat{\+\beta}_{(i)}-\widehat{\+\beta})$. If we wanted to reduce this to a single scalar summary statistic, we could employ the Euclidean inner product to measure the (squared) magnitude of the overall displacement: 
\begin{equation}
\label{euclidean}
 (\widehat{\+\beta}_{(i)}-\widehat{\+\beta})^T(\widehat{\+\beta}_{(i)}-\widehat{\+\beta})
 \end{equation}
Cook's distance is similar in spirit, but employs a more general inner product; in particular, it satisfies
\begin{equation}
\label{cooks}
c_i \propto (\widehat{\+\beta}_{(i)}-\widehat{\+\beta})^T \; Cov(\widehat{\+\beta})^{-1} \; (\widehat{\+\beta}_{(i)}-\widehat{\+\beta}) 
 \end{equation}
where $Cov(\widehat{\+\beta})$ is the covariance matrix of the fitted regression weights. 

To understand the derivation, recall that the estimator $\widehat{\+\beta}$ is asympotically distributed as a multivariate normal distribution:
\begin{equation}
\label{distribution}
\widehat{\+\beta} \stackrel{\bigcdot}{\sim} N\big(\+\beta, Cov(\widehat{\+\beta} ) \big)
\end{equation}
This implies, since $Cov(\+\beta)=(\+X^T\+X)^{-1} \sigma^2$, and using standard facts about multivariate normals, that
\begin{equation}
\label{chisquared}
(\widehat{\+\beta}-\+\beta)^T \; \df{\+X^T\+X}{\sigma^2}\; (\widehat{\+\beta}-\+\beta)  \stackrel{\bigcdot}{\sim} \chi^2_p 
 \end{equation}

Now we switch our point of view, making substitutions\footnote{In Cook's original formulation~\cite{cook}, these substitutions were made only after estimating the unknown variance, $\sigma^2$, with the sample variance, $s^2$, obtaining a ratio of independent chi-squared random variables divided by their respective degrees of freedom, and thereby deriving an F-statistic: $ \df{(\widehat{\+\beta}-\+\beta)^T \; \+X^T\+X \; (\widehat{\+\beta}-\+\beta)}{ps^2}  \stackrel{\bigcdot}{\sim} F(p,n-p)$.  However, as mentioned, the constant denominator is not relevant for making comparisons across samples; thus, it has become more common (see, e.g. ~\cite{pregibon},~\cite{hosmer}) to use the form in Equation~\ref{cooks3}.} about what is known,  $\+\beta \leftarrow \widehat{\+\beta}$,  and what is being estimated, $\widehat{\+\beta} \leftarrow \widehat{\+\beta}_{(i)}$.   Throwing away the constant denominator, which is not relevant for making comparisons across samples, we obtain: 
 \begin{equation}
 \label{cooks3}
c_i = (\widehat{\+\beta}_{(i)}-\widehat{\+\beta})^T \; \+X^T\+X \; (\widehat{\+\beta}_{(i)}-\widehat{\+\beta})
 \end{equation}
 which satisfies Equation~\ref{cooks}.
 
Thus, Cook's distance effectively imposes the model 
\begin{equation}
\label{imagined}
\widehat{\+\beta}_{(i)} \sim N\bigg(\widehat{\+\beta}, \; Cov(\widehat{\+\beta})\bigg)
\end{equation}
More precisely, Cook's distance is proportional to the log kernel of the multivariate normal density $N\big(\widehat{\+\beta}, \; Cov(\widehat{\+\beta})\big)$ evaluated at the perturbed regression weights, $\widehat{\+\beta}_{(i)}$.

\subsubsection{Interpretation} \label{strange}
In a sense, this distributional assumption for $\widehat{\+\beta}_{(i)}$  is duplicitous.  After the substitutions made in Section~\ref{cooks_stat}, the probability distribution assigned to $\widehat{\+\beta}_{(i)}$ is no longer correct.    As just one example, the frequentist derivation of the quantity in Equation~\ref{chisquared} assumes predictor variables $\+X$ are fixed (rather than random variables), but case deletion alters $\+X$.  
However, Cook constructed his $c_i$ statistic not for the purpose of hypothesis testing or creating confidence intervals using $\widehat{\+\beta}_{(i)}$, but as a sheer measure of \emph{distance} between the original and perturbed regression weights.   As seen by Equation~\ref{imagined}, Cook's Distance monotonically increases as the likelihood of observing $\widehat{\+\beta}_{(i)}$ under the model of Equation~\ref{imagined} decreases.  In other words, a sample is influential if the model fit \emph{without} that sample has regression weights that look unusual based on knowledge obtained from the model fit \emph{with} that sample.  In particular, to determine what counts as an ``unusual" vector of regression weights, Cook's distance incorporates information about the covariance structure of $\widehat{\+\beta}$, yielding a substantial advantage over the simpler Euclidean inner product from Equation~\ref{euclidean}.  This produces nice properties; for example, (a) Individually speaking, some features have inherently more uncertain (i.e., higher variance) regression weights than others.  Thus, if a sample perturbs these high variance regression weights, it would be considered less influential than a sample which perturbs lower variance regression weights.  (b) Collectively speaking, holding magnitude of change constant, certain distributions of variation across multiple regression weights are more expected (i.e., more likely) than others.  An influential sample perturbs the regression weights in unlikely directions.

 \begin{figure}
\centering
\includegraphics[height=3.4in]{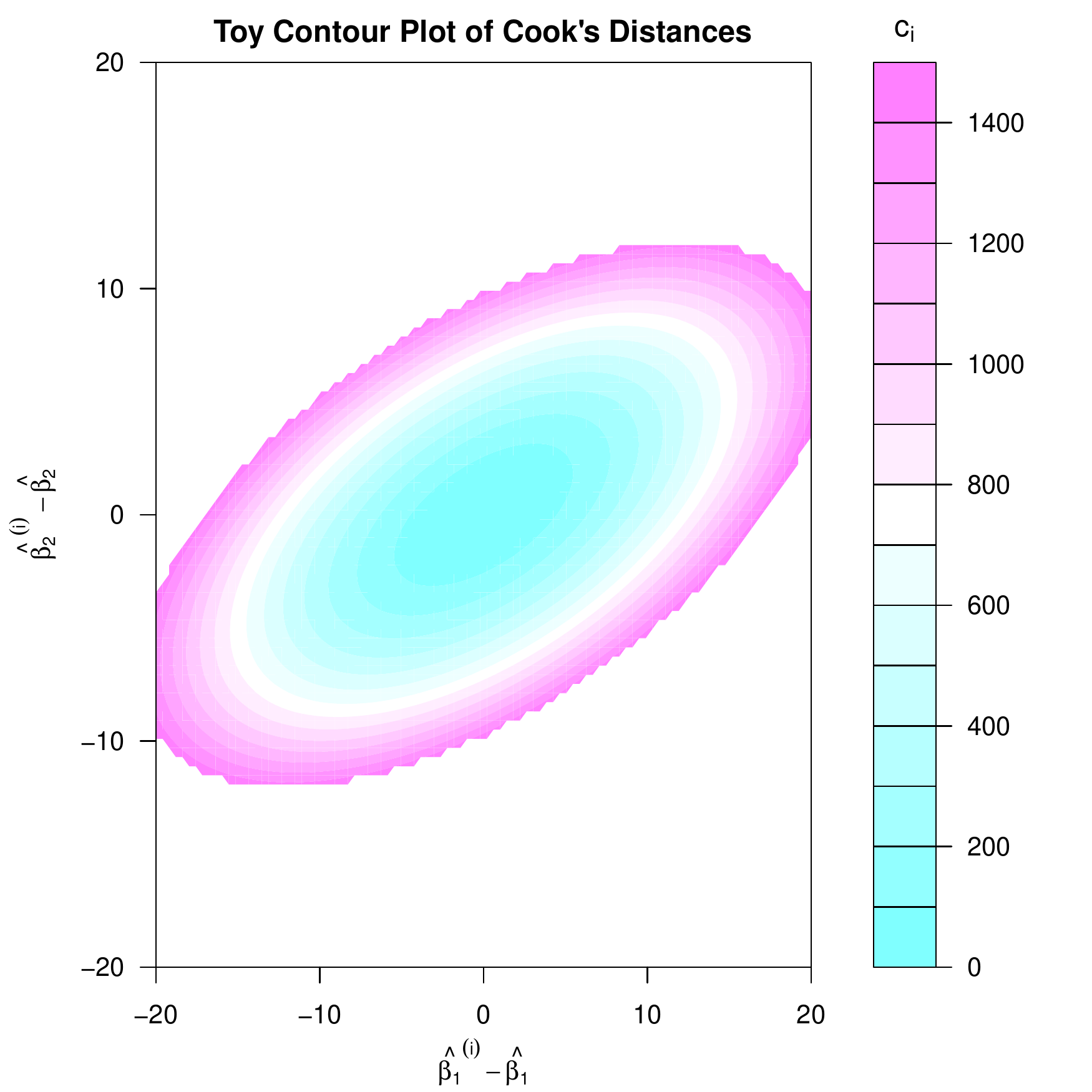}
\caption{ \emph{Illustration of Cook's Distances.}  If case deletion produces a new regression vector $\widehat{\+\beta}_{(i)}$ which is not likely given the distributional assumption $\widehat{\+\beta}_{(i)} \sim N\big(\widehat{\+\beta}, \; Cov(\widehat{\+\beta})\big)$, then the sample has a large Cook's distance. }
\label{toycontour}
\end{figure}

Figure~\ref{toycontour} provides a toy example illustrating these properties, which hold because Cook's Distance depends upon the estimated covariance structure of the regression weights.  
For the purposes of the toy example, we construct a bivariate feature space where the empirical covariance matrix for the regression weights is preset to $Cov({\widehat{\+\beta}}):= \begin{bmatrix} .3 & .1 \\ .1 & .1 \end{bmatrix}$. 
We see here that the $\widehat{\+\beta}_{(i)}$'s on the elliptical isocontours of the estimated probability distribution of Equation~\ref{distribution} have equivalent Cook's distances.     In other words, two samples which cause equally-sized model displacements but in different directions will generally not be considered, by Cook's metric, equally influential.   The main point is that Cook's distance determines a sample's influence not simply by measuring the \emph{size} of its perturbation on the regression weights, but also by evaluating the  \emph{direction} of perturbation.  In sum, \emph{influential samples cause not merely large -- but also strange -- perturbations of regression weights.} 
\subsubsection{\sloppy Computational Form} 
\label{computational_form}

Since fitting case deletions requires an iterative procedure, computing $\widehat{\+\beta}_{(i)}$ for each sample would be difficult.  This is especially true in the large-scale case and with a regularized generalized linear model such as the logistic lasso; indeed, in the example we consider here, it would require fitting over 2 million large-sample high-dimensional models, each one of which might take on the order of an hour.  However, Cook (1977,1979) developed a way to estimate the quantity in Equation~\ref{cooks3} directly from the Hat matrix. In particular, Cook~\cite{cook} showed that
\begin{equation}
\label{cooks_trick}
 \widehat{\+\beta} - \widehat{\+\beta}_{(i)} = \df{(\+X^T\+X)^{-1}\+x_i}{(1-h_i)} (\+y_i-\widehat{\+y}_i)
 \end{equation} 

Using this equation, we can determine $ (\widehat{\+\beta} - \widehat{\+\beta}_{(i)})$, the change in regression weights caused by case deletion, as a function of already known quantities (meaning that it would not be necessary to rerun the regression).     Substituting this trick  into the definition for Cook's Distance given in Equation~\ref{cooks3}, we obtain a computational formula for Cook's Distance:

\begin{equation}
\label{computational_form}
c_i= \widehat{r_{i}^2} \; \df{ h_i}{(1-h_i)} 
 \end{equation} 
 
 
\subsubsection{\sloppy Decomposition} 
\label{leverage_times_discrepancy}

The computational form for Cook's distance (Equation~\ref{computational_form}) enables a decomposition of influence into component concepts.  Note that the first term in the product, $\widehat{r_{i}^2} $, quantifies the extent to which the sample is ``discrepant." A sample is discrepant (or a regression outlier) if it has an unusual $y_i$ observation given its $\+x_i$ predictors and the fitted model.  The second term in the product, $\df{ h_i}{(1-h_i)}$, describes whether the sample has ``leverage."  In Section~\ref{leverage_primer}, we noted that a sample has high leverage if it has an extreme value of the vector $\+x_i$.   With these definitions in mind, we can immediately see, embedded within the computational form of Cook's Distance (Equation~\ref{computational_form}), a common moniker for conceptualizing influence:  
\[ \text{Influence} = \text{discrepancy} \times  \text{leverage}  \]



%


\subsection{Generalized Cook's Distance: Extending Influence to Generalized Linear Models (GLMs)} 
\subsubsection{Construction} \label{extension_glms}
 
Pregibon~\cite{pregibon} developed an approximation for Cook's distance in the case of generalized linear models (which can handle observations whose error distributions are not normal, but rather binomial, Poisson, beta, etc.).  To see how, consider that the so-called \quotes{normal equations} used to solve for $\widehat{\+\beta}$, $\+X^T\widehat{\+r} = \+0$, are nonlinear in $\widehat{\+\beta}$, and so iterative methods are required to solve them.   A common approach is to apply Newton-Raphson technique, which, after some algebra, leads to the iterative expression
\begin{equation}
\label{irls}
\widehat{\+\beta}^{t+1} = (\+X^T \+V^t  \+X)^{-1} \+X^T \+V^t \+z^t
\end{equation}
where $\+V^t = diag\big(\widehat{\+p}_i^t (1- \widehat{\+p}_i^t )\big)$ for logistic regression (and where the matrix can be similarly derived from the exponential family representation of other probability distributions from the set of generalized linear models) and where 
\begin{equation}
\label{z}
\+z^t = \+X\widehat{\+\beta}^t + (\+V^{t})^{-1}\widehat{\+r}^t
\end{equation}

The iterative procedure in Equation~\ref{irls} is known as \emph{iterative re-weighted least squares} (IRLS). In particular, upon convergence (and so where we have obtained $\+V:= \lim_{t \to \infty} \+V^{t}$), the maximum likelihood estimator for $\+\beta$ can then be expressed as the solution of a weighted least squares problem:
\begin{equation}
\label{wls}
\widehat{\+\beta} = (\+X^T \+V  \+X)^{-1} \+X^T \+V \+z
\end{equation}
In other words, the same regression weights would be obtained if we had performed a simple linear regression of psuedo-observations $\+V^{1/2}\+z$ on psuedo-predictors $\+V^{1/2}\+X$. 

Pursuing this observation, Pregibon~\cite{pregibon} obtained generalized regression diagnotics. In particular, we can let $t$ go to infinity in Equation~\ref{z} , premultiply the result by $\+V^{1/2}$, and apply the following substitutions: 
\begin{align*}
\+X & \leftarrow \+X^* = \+V^{1/2} \+X  \numberthis \label{substitution}  \\
\+y & \leftarrow \+y^* = \+V^{1/2} \+z   \\
\widehat{\+r} & \leftarrow  \widehat{\+r}^* = \+V^{-1/2} \; \widehat{\+r}
\end{align*}
Using these substitutions, we can transform the original predictors, observations, and residuals into their psuedo-versions, and thereby obtain the following linearized representation (available only after convergence) of the generalized linear model:
\begin{equation}
\label{linearized}
\+y^*=  \+X^*\widehat{\+\beta} + \widehat{\+r}^* 
 \end{equation}
This representation renders all the model diagnostics from linear regression available to generalized linear regressions. For instance, the hat matrix becomes:

\begin{align*}
 \+H^* &= \+X^* (\+X^{*T} \+X^*)^{-1} \+X^{*T} \numberthis \label{newhat} \\
 &= \+V^{1/2}\+X (\+X^T \+V \+X)^{-1} \+X^T \+V^{1/2} \\
 \end{align*}
And the generalized version of Cook's distance becomes:

\begin{equation}
\label{generalized}
c_i= (\widehat{r_{i}}^{*})^2 \; \df{ h^*_i}{(1-h^*_i)} 
 \end{equation} 
 
 \subsubsection{Computing $\+V$, the IRLS weight matrix for Generalized Cook's Distance} \label{computing_irls}
 
 Generalized linear models (normal, logistic, multinomial, Poisson, beta, etc. regressions) model response variables $y_i$ through probability distributions from the ``exponential family."  These distributions have probability densities for observations that are dependent upon location and scaling parameters, $\theta$ and $\phi$, respectively, and can be expressed in the following form (for a single observation, $y_i$):
 \[f(y_i; \theta,\phi) = \exp \bigg\{ \df{y_i \theta_i - b(\theta_i)}{a(\phi)} + c(y_i,\theta)\bigg\}  \]
 where typically $a(\phi) = \phi/\rho_i$.  
 
The vector of predictor variables for a given sample, $\+x_i$, are incorporated by modeling the expected value of the response, $\mathbb{E}Y_i := \mu_i$, as a function of the so-called ``linear predictor" $\eta_i = \+x_i^T \hat{\+\beta}$ through a link function, $g(\cdot)$, via $\mu_i = g^{-1}(\eta_i)$.   (For example, in logistic regression, $g(\cdot)$ is the logit function). 

For any generalized linear model, we can express the IRLS weight matrix, $\+V \in \R^{n \times n} = diag(v_i)$, through the following relation:
\begin{equation}
\label{Vmatrix}
v_i = \df{\rho_i}{b''(\theta_i)\bigg(\df{\partial \eta_i}{\partial \mu_i}\bigg)^2}
\end{equation}

For instance, for logistic regression, $v_i = \widehat{p}_i (1-\widehat{p}_i)$, and for linear regression, $v_i = 1$. 
 
\section{\sloppy ``Influence Sketching": Measuring Influence in Large Scale Regressions}

\subsection{Scalability issues with the classical method} 

For large scale regressions, the computation of the hat matrix is infeasible.  In Equation~\ref{hat_def}, we see that the hat matrix requires computation of $(\+X^T\+X)^{-1}$, which can be infeasible for high dimensional datasets with many samples.  Thus, it is not immediately clear how to identify unusually influential samples in this setting. 

 \subsection{Random Projections} \label{rp}
 
Randomized algorithms, such as random projections~\cite{mahoney}, have become a widely used approach for handling very large matrix problems.   Given a $n \times p$ matrix $\+X$, which we interpret as $n$ samples in $p$ dimensional-space, a random projection involves post-multiplying $\+X$ by a $p \times k$ \emph{random projection matrix} $\+\Omega$.    The resulting randomly projected dataset, $\+Y = \+X \+\Omega$ has only $k$ dimensions instead of the original $p$ dimensions, thereby speeding up computations dramatically.   The Johnson-Lindenstrauss (JL) embedding theorem~\cite{dasgupta} can be applied to show that this procedure approximately preserves pairwise distances between the $n$ samples so long as $k$ is chosen on the order of $\log{n}$.    

There are a number of ways to construct such a matrix.  For instance, a \emph{Gaussian random projection} has entries which are i.i.d. Gaussian $N(0,1)$ random variables.  A \emph{very sparse random projection} involves constructing $\+\Omega$ with entries in $\{-1,0,1\}$ with probabilities $\{ \df{1}{2\sqrt{p}}, 1-\df{1}{\sqrt{p}}, \df{1}{2\sqrt{p}}\}$.  This construction is especially useful for very large problems, as it  produces a $\sqrt{p}$-fold speedup with little loss in accuracy~\cite{hastie}.  

Of crucial interest to us here is that randomly projected datasets have approximately the same column space (or ``range") as the original dataset, so long as $k$ is chosen large enough to at least approximate the effective dimensionality, or numerical rank, of the dataset~\cite{mahoney},~\cite{halko}.  The intuition is as follows:  if $\+\Omega_i$ is the $i$th column of $\+\Omega$, then $\+X\+\Omega_i$ will obviously lie in the range of $\+X$.  But by the random sampling, the columns $\+\Omega_i$ are very likely to be linearly independent (although possibly poorly conditioned), and thus to have a $k$-dimensional range. 
We summarize this fact as $ \mathcal{C_{\+X}} \approx \mathcal{C_{\+Y}}$, and will use it in Section~\ref{justification} to justify the use of random projections in an algorithm for measuring the influence of samples in large-scale regressions.   




\subsection{Proposed Solution}

In Algorithm~\ref{sample_influence} (``Influence Sketching"), we describe how to calculate approximate sample influence scores for large-scale regressions from the GLM family.   The algorithm embeds random projections inside the hat matrix by randomly projecting the pseudo-predictors, $\+X^{*} = \+X\+V^{1/2}$, defined in Section~\ref{extension_glms}.  We define the randomly projected psuedo-predictors as $\+Y^{*}=\+X^{*} \+\Omega = \+V^{1/2}\+X\+\Omega$, where $\+\Omega$ is a random projection matrix.  In comparison to the generalized hat matrix of Equation~\ref{newhat}, which we now more explicitly denote as $\+H^*_X$, the generalized hat matrix for large-scale regressions is:
\begin{align*}
 \+H^{*}_Y &=  \+V^{1/2} \+Y (\+Y^T \+V \+Y)^{-1} \+Y^T \+V^{1/2}  \numberthis \label{hat_rp} \\
 &= \+V^{1/2}\+X\+\Omega (\+\Omega^T\+X^T \+V \+X\+\Omega)^{-1} \+\Omega^T \+X^T \+V^{1/2} \\
 \end{align*}

The algorithm can find influential samples from large-scale regressions of various types from the GLM family by appropriate choice of the matrix $\+V$; for instance, for linear regression we have $\+V=\+I_n$, the identity matrix, and for logistic regression we have $\+V =diag(\widehat{p}_i (1-\widehat{p}_i))$, where $\widehat{p}_i$ is the fitted probability that the sample takes on a response value of 1.   Note that for logistic regression, the expression $\widehat{\mathbb{E}}[\+y|\+X]$ in the residual formula simply evaluates to $\widehat{p}$, the vector of fitted probabilities that the samples have binary responses of 1 given their predictor variables.

\begin{algorithm}
\caption{{\bf (Influence Sketching):} Calculating sample influence for large scale regressions}
\label{sample_influence}
\begin{algorithmic}[1]
\Statex {\bf Data}  A dataset $\+X \in \R^{n \times p}$, a random projection matrix $\+\Omega \in \R^{p \times k}$, and a regression model chosen from the family of generalized linear models. 
\Statex {\bf Result} A vector $\+c \in \R^{n}$ quantifying the influence of each sample on the model fit. 
\vspace{.2in}
\State From the regression model, obtain the fitted observations $\widehat{\mathbb{E}}[\+y|\+X]$ and the converged IRLS (iteratively reweighted least squares) diagonal weight matrix,  $\+V \in \R^{n \times n}$, where the latter can be constructed using a lookup table or the relationship in Equation~\ref{Vmatrix}.
\State Compute the re-weighted residuals $\widehat{\+r^*}=\+V^{-1/2} (\+y -\widehat{\mathbb{E}}[\+y|\+X])$
\State Form re-weighted randomly projected data: $\+Z  \in \R^{n \times k}: \+Z=\+V^{1/2} \+Y$, where $\+Y=\+X\+\Omega$ is the original dataset randomly projected to $k$ dimensions.  \label{new_data}
\State Form the inverse-covariance matrix of the re-weighted randomly projected data : $\+W \in \R^{k \times k}:  \+W = (\+Z^T \+Z)^{-1}$ \label{inverse_cov}
\State Get the leverage values as the diagonal elements of the generalized hat matrix, $h_i^* \in \R : h_i^*= \+z_i ^T \+W \+z_i$ \label{leverage}
\State Compute the approximate influence (or generalized Cook's distance) scores as $c_i =(\widehat{r^*}_i)^2 \df{h_i^*}{(1-h_i^*)}$
\end{algorithmic}
\end{algorithm}

\subsection{Justification} \label{justification}

The intuition behind the ``influence sketching" algorithm is as follows: recall that the role of the hat matrix, $\+H$, is to project responses $\+y$ onto the column space of $\+X$.  (The same argument holds for the generalized hat matrix, $\+H^*$, which projects pseudo-responses $\+y^*$ onto the column space of the pseudo-predictors, $\+X^*$.)   Since random projections preserve the column space of the dataset, it should be possible to embed random projections inside the construction of the hat matrix to obtain approximate sample influence scores.   

In detail, consider that the hat matrix $\+H^*_{X}:=\+X^*(\+X^{*T}\+X^*)^{-1}\+X^{*T}$ is a projection matrix which projects vectors onto the column space of $\+X^*$. 
That is, if we denote the column space of $\+X^*$ by $\mathcal{C}_{X^*}$, and the Eulidean projection onto $\mathcal{C}$ by $\Pi_{\mathcal{C}}$, then we have $\Pi_{\mathcal{C}} = \+X^*(\+X^{*T}\+X^*)^{-1}\+X^{*T}$, in the sense that for any vector $\+v \in \R^n$, we have:
\begin{align*}
\Pi_{\mathcal{C}_{X^*}}(\+v) & := argmin_{\+x \in \mathcal{C}_{X^*}} ||  \+x- \+v ||_2 \\
& = \+X^*(\+X^{*T}\+X^*)^{-1}\+X^{*T}\+v = \+H^*_X\+v 
\end{align*}

Now by the Hilbert Space Projection Theorem, and since $\R^n$ is a Hilbert Space, we have
\begin{align*}
&\small \mbox{\emph{a projection onto any closed subspace of} }\R^n  \\
&\mbox{\emph{produces a unique element}} \normalsize \numberthis \label{hilbert} 
\end{align*}

That is, the image of the transformation will not be affected by redescriptions of the set $\mathcal{C}$.  Thus, if two matrices $\+Y^*$ and $\+X^*$ have identical column spaces, then their hat matrices are equal, because
\begin{align*}
 \+H^*_Y & = \+Y^*(\+Y^{*T}\+Y^*)^{-1}\+Y^{*T} \\
 &=  \Pi_{\mathcal{C}_{\+Y^*}} \stackrel{(\ref{hilbert})}{=} \Pi_{\mathcal{C}_{\+X^*}}  \\
&=  \+X^*(\+X^{*T}\+X^*)^{-1}\+X^{*T}  = \+H^*_{X}  
 \end{align*}
 
 Now, as discussed in Section~\ref{rp}, from random projection theory, we know that $ \mathcal{C_{\+X^*}} \approx \mathcal{C_{\+Y^*}}$  (That is, if $\+\Omega \in \R^{p \times k}$ is a random projection matrix, then the column space of $\+Y^* \in \R^{n \times k} : \+Y^* = \+X^* \+\Omega$ is approximately equal to the column space of $\+X^*$, so long as $k$ is sufficiently large).  Thus, so long as $k$ is sufficiently large, $\+H_Y^* \approx \+H_X^*$.

\subsection{Computational complexity} \label{computation}
We now investigate the computational complexity of Algorithm~\ref{sample_influence},  assuming the relatively unlikely worst case scenario that the dataset and random projection matrices are dense.   In practice, however, large scale data-sets are sparse, and the random projection can be made sparse as well, dramatically improving algorithmic performance.    
The formation of the proxy dataset in Step~\ref{new_data} is $\mathcal{O}(npk)$.   This is the cost of the matrix multiplications (recalling that the $\+V^{1/2}$ matrix is diagonal).    The formation of the inverse covariance matrix of the proxy dataset in Step~\ref{inverse_cov} is $\mathcal{O}(nk^2 + k^3)$.  This is the cost of another matrix multiplication followed by a matrix inverse.   Finally, the computation of the leverage in Step~\ref{leverage} is $\mathcal{O}(nk^2)$ considered as a for loop over matrix multiplications.    

Thus, the algorithm has (worst-case) complexity $\mathcal{O}(npk+nk^2+k^3)$. We can re-express this in terms of $n$ and $p$ only by considering that, by the (JL) embedding lemma~\cite{achlioptas}, distances between samples can be well-preserved to a particular pre-specified error tolerance when $k$ increases on the order of $\log(n)$.  Since for most applications $\log^3(n) < \min\{p,n\}$, the algorithm has an approximate worst-case complexity of $\mathcal{O}(np \;log(n)).$



\subsection{Large Scale Regularized GLMs} \label{ll_extend}
Strictly speaking, the argument above does not apply in a straightforward manner to large scale regularized GLMs, such as lasso logistic regression.  The argument so far depends upon the fact that GLMs can be solved by IRLS (and therefore re-represented, after convergence, as a linear regression with re-defined predictors and responses). However, in this setting, the logistic lasso~\cite{lee} is solved by incorporating a \emph{least angle regression} (LARS) algorithm at each step of the IRLS algorithm (which amounts to solving a linear lasso, rather than a linear regression, at each iteration).    Whether or not the arguments above, which justify influence sketching, would extend to logistic lasso is unclear, as are the appropriate adjustments (if any) to the influence sketching algorithm to handle the case of regularization. We leave this as a direction for future research.   For now, we just imagine that the regression weights produced by a regularized regression were produced by an unregularized regression, even though this approach could theoretically produce grave distortions in assessing influence, since there would be  inappropriate estimates for the quantity $Cov(\widehat{\+\beta})$ which lies at the heart of Cook's Distance (see Section~\ref{cooks_stat}).


\section{Experiment}

\subsection{Motivation}
In the experiment, we address two primary questions:  

\begin{enumerate}
\item \emph{Can influence sketching help to prioritize samples for human expert attention?} For example, a priority queue could be constructed on nominal model miscategorizations to search for influential mislabeled samples, whose labels could be flipped~\cite{frenay}.   For priority queuing, we would like three conditions to hold:
    \begin{enumerate}
         \item The \emph{rare event condition} refers to the fact that the distribution of influence sketch scores should be heavily right-skewed, meaning that a relatively small number of samples accounts for a relatively large amount of influence on the fitted model.  
    \item \emph{Validity} refers to the fact that the influence sketching algorithm, despite using randomization, should track true (non-approximated) influence scores.   If so, the samples deemed influential by the influence sketching algorithm should indeed have disproportionately large influence on the fitted regression betas, and omitting such samples from training (but not testing) should reduce predictive accuracy.  
    \item \emph{Reliability} refers to the fact that the ranking of samples by influence should not vary much from random projection matrix to random projection matrix (assuming sufficiently large choice of latent dimensonality, $k$.)   

 \end{enumerate}
\item \emph{Are samples with large influence sketch scores particularly likely to be mislabeled?}  In the framing of the previous question, the priority queue simply focuses the search for mislabeled samples to the  candidates which matter most to the model.  In this way, the priority queue would be valuable even if the proportion of miscategorized samples which are mislabeled does not covary with influence scores. However, when miscategorized samples are also influential, they may in fact be particularly likely to be mislabeled, in which case prioritizing by influence additionally enhances the probability of discovering labeling errors.   The argument for this hypothesis, following the moniker that \emph{influence} = \emph{discrepancy}  $\times$  \emph{leverage} (see Section ~\ref{leverage_times_discrepancy}), is that such samples are not only miscategorized, but also have large model residuals, and are located in strange regions of feature space (at least with regards to the pseudo-dataset $\+X^*$). Both conditions are suggestive of label noise: large model residuals suggest that model may be struggling to fit an incorrect label, and high leverage (reflecting that the sample is located in an usual location of feature space) suggests that the labeling mechanism may be particularly error-prone (e.g., in the case of cybersecurity, it's possible that signature-based detection would struggle to accurate identify such pieces of malware).  
\end{enumerate}

\subsection{Data} \label{Data}

Training data are $N_{\text{train}}$=2,342,274  \emph{Portable Executable} (PE) files.\footnote{The Portable Executable (PE) file format is the standard executable file format used in Windows and Windows-like operating systems.}  These files were labeled as either \quotes{clean} or \quotes{malicious} by a automatic labeling algorithm based on signature-based detection (see, e.g.~\cite{kantchelian}).  Using this method, 1,374,226 (58.67\%) of the files were determined to be malicious.  The remainder of the files were determined to be clean.    The "malware" category contained different types of malicious software (e.g., computer viruses, Trojan horses, spyware, ransomware, and adware).   An additional $N_{\text{test}}$=562,066 files (62.35\% malware) were obtained for the test set.  From each portable executable file, we extracted $P=98,450$ features previously determined to be relevant to whether a file was malicious or not.  The features were sparse, and mixed continuous and binary.  The density level was approximately .0224.  These features were also used in~\cite{wojnowicz}. 

Due to the presence of label noise, we obtained software vendor reputational scores for a subset ($150,751$, 15.57\%) of the clean samples from the training set.  Clean samples were analyzed because the labels were obtained by an algorithm based on signature-based detection, which is known to have  low false positive but high miss rates (see, e.g.~\cite{kantchelian}).  Of these samples, 47,193 (31.31\%) were rated as coming from ``highly trusted" sources (such as Microsoft).  When a nominally good sample was developed by a highly trusted source, we say that it has a \emph{highly trusted label.}

\subsection{Methods}  \label{Methods}

To create the influence sketches, we chose a very sparse random projection matrix (see Section~\ref{rp}). This type of projection matrix was chosen to partially preserve the sparsity in our dataset and to speed computation time.  As recommended by \cite{hastie}, the density of the random projection matrix was set to $1/\sqrt{P}\approx.003$. We set $K=5000$ based on the results of~\cite{wojnowicz}.    Obtaining influence scores for all 2.3 million samples took about an hour and a half using non-optimized scripts written in the Julia programming language and executed on a single r3.8x instance (with 32 workers for parallel processing) in Amazon's EC2 system.  

To validate the technique, we applied two types of regression models for binary classification: a logistic regression and lasso logistic regression (i.e. L1-regularized logistic regression).  Because constructing the optimal lasso logistic regression model was not directly relevant to the goals of this paper, the lasso logistic regression was made parameter free by constructing the cost function such that equal weight was placed on the negative log likelihood and the L1-regularization term.  We fit each type of regression model to the dataset, and then we performed an ``ablation experiment" where we deleted fixed numbers of samples  from the training set.  These samples were either influential samples or randomly selected samples.  We then retrained the regression models so that we could observe the effect on predictive performance on the hold-out test set.  The number of files to ablate, $N_{\text{ablate}}$, was set to: (a) 23, (b) 23,000, (c) 115,000, and (d) 230,000,  constituting approximately .001\%, 1\%, 5\%, and 10\% of the training corpus, respectively.   We compared the effect on the model of ablating the $N_{\text{ablate}}$ most influential samples, vs. $N_{\text{ablate}}$  randomly selected files.   

For the model training error analysis, performed on the full training set, the influential sample set was defined as the top 23,000 ($\approx 1\%$) of influential samples.  For the label trustworthiness analysis of the 150,751 nominally good samples for which label trustworthiness judgments were available, the ``very high" influence group was defined as the .1\% most influential samples (i.e. ranks 1-151), the ``high" influence group as the next .1\% most influential such samples (i.e. ranks 152-253), and the ``not high" influence group as the remaining samples.

\subsection{Results and Discussion }

\subsubsection{Can influence sketching help to prioritize samples for human expert attention?}  

\begin{figure}
\centering
\includegraphics[height=2.3
in]{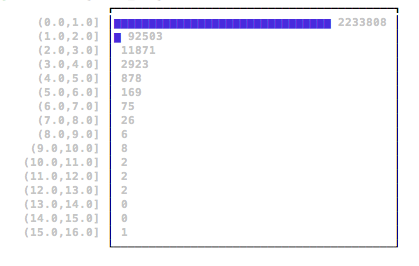}
\caption{ \emph{Distribution of Influence Sketch Scores.}  The distribution of influence sketch scores is heavily right-skewed.}
\label{histogram}
\end{figure}

We begin by addressing the {\bf rare event condition} for influence sketching.  The distribution of influence sketch scores for the full training set is shown in Figure~\ref{histogram}.  The distribution of influence for these portable executable files is heavily right-skewed, suggesting that a relatively small percentage of samples is much more influential than the other samples.  For example, 6.3\% of the total influence (the sum of the influence scores) were achieved by a mere 0.1\% of the samples.  This suggests that we can obtain an asymmetrically large payoff from investment by restricting any manual analysis to influential samples.  

Next, we address the {\bf validity} of the influence sketching algorithm.   Samples with large influence sketch scores are rare, but is the influence sketching algorithm truly finding influential samples, despite the approximations from the random projection matrix?  We find that the displacement in the beta vector, measured in terms of the L1-norm (or sum of the absolute values of its elements) is substantially larger when influential samples are removed than when random samples are removed. For instance, for the logistic lasso, the norm $||\Delta \widehat{\+\beta}||_1$ when ablating random vs. influential samples was (4486.07 vs. 4646.35) for ablations of size 23, (4,979.38 vs. 30,504.26) for ablations of size 23,000, and (17,048.51 vs 79,026.27)  for ablations of size 230,000.  Although, as discussed in Section~\ref{strange}, displacements in the \emph{magnitude} of $\widehat{\+\beta}$ is only part of what influence measures (directionality matters as well), this finding corroborates the notion that influence sketching is finding influential samples.  

To further emphasize this point, we ask: do the samples flagged as influential by the influence sketching algorithm actually impact the model's downstream predictive performance on a hold-out test set?  A skeptic might argue, based on the moniker that $\emph{influence} = \emph{leverage} \times \emph{discrepancy}$, that influential samples could conceivably hugely impact the regression betas without having a large impact on the predictive performance.  The argument in this case would be that the regression betas might shift dramatically in order to handle the samples in extreme locations of feature space (equivalent to dragging the far-right point, and therefore the fitted regression line, downward in Figure~\ref{leverage}), but that this large shift in regression weights would have a relatively minor influence on the fitted or predicted values for most samples, which lie in more populated regions of feature space (indeed, even if the regression line were dragged down to be horizontal, the fitted values for the large cloud of points in Figure~\ref{leverage} would remain about the same).  

\begin{figure}
\centering
\includegraphics[height=3in]{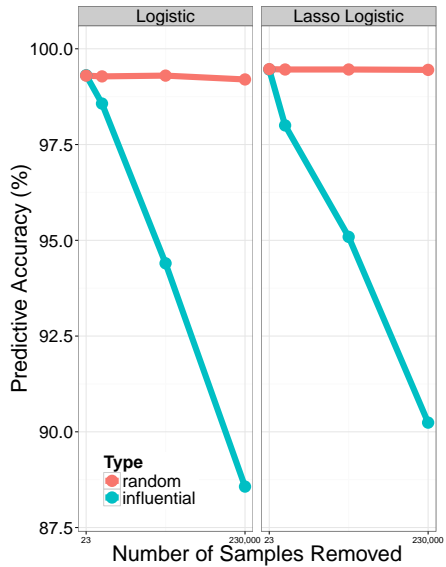}
\caption{ \emph{Influential samples, as determined by influence sketching, disproportionately drive model performance.} The plot shows the predictive performance on a hold-out test set after removing up to $\approx$10\% of the training set of $\approx$2.3 million samples, where removed samples are either influential or random.  The model on the left is a standard (unregularized) logistic regression, whereas the model on the right is a L1-regularized (``lasso") logistic regression.}
\label{LogRegResults}
\end{figure}

In contrast to this logic, we find that the influential samples identified by the influence sketching algorithm are exceptionally impactful on a large-scale regression model's predictive performance.   Figure~\ref{LogRegResults} shows results of the ablating experiment, where we find that removing influential samples from the training set substantially deteriorates model predictive performance on an unaltered hold-out test set.  The figure shows that removing up to 10\% of samples from the training corpus, so long as they are selected randomly, has no discernible effect on predictive performance.   For example, the full logistic lasso model with all samples obtains 99.47\% predictive accuracy, whereas randomly deleting approximately 10\% of the training samples reduces accuracy only slightly to 99.45\%. In contrast, deleting the same number of influential samples (the far right tail of Figure~\ref{histogram}) from training reduces predictive accuracy all the way down to 90.24\%.  Clearly, the model is disproportionately strongly affected by influential samples, which can be a problem if there are systematic labeling errors in the sample set.\footnote{We note that if systematic labeling errors are uniformly distributed across training and test sets, as they should be by random assignment, then removing mislabeled samples from the training set (or relabeling them), without analogously adjusting the test set, may \emph{nominally} deteriorate model performance, while improving \emph{actual} model performance with regards to accurate labels.  However, we do not believe that all influential samples are mislabeled.}   
Indeed, in Section~\ref{results2}, we demonstrate that influential samples are, in fact, particularly likely to be mislabeled.  However, the main point here is  simply about the validity of influence sketching algorithm:  manipulating the training set by removing samples with high influence sketch scores significantly altered the training distribution with respect to model learning, but the same did not occur when removing random samples.  Thus, the influence sketching algorithm is indeed finding influential samples. 

\begin{figure}
\centering
\includegraphics[height=2.5in]{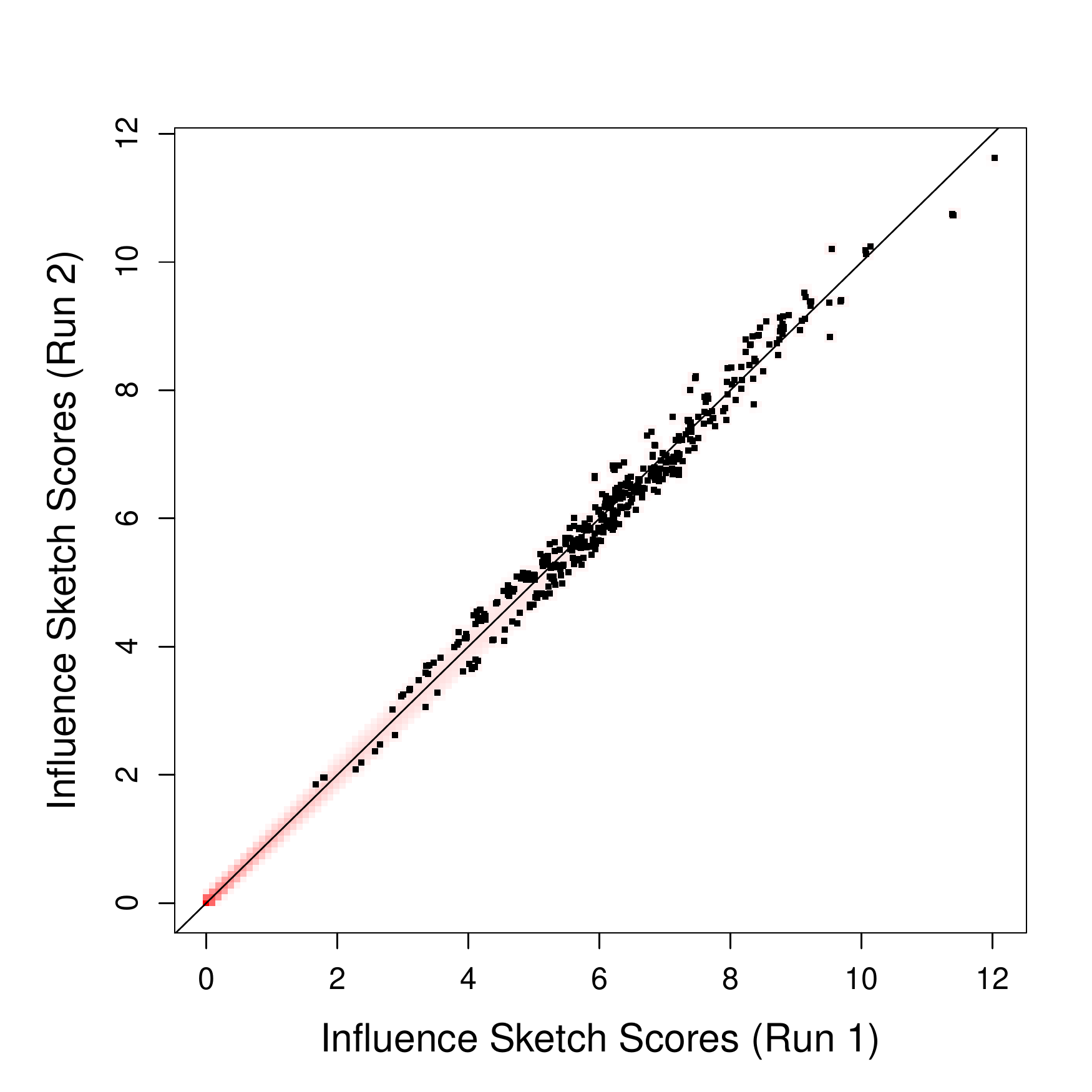}
\caption{ \emph{Reliability of influence sketch scores across two runs.}}
\label{reliability}
\end{figure}

Finally, we address the {\bf reliability} of the influence sketching algorithm.   We ran the influence sketching algorithm twice, with two independently sampled random projection matrices of the type described in Section~\ref{Methods}.  Despite the approximate nature of influence sketching due to the random projection, we obtain very similar influence scores (and nearly identical ranking of samples according to influence), as expected by the justification for influence sketching provided in Section~\ref{justification}.  Indeed, the correlation between influence sketch scores across runs was very high (r=.9995).  In Figure~\ref{reliability}, we display a smoothed color density representation of a scatterplot, obtained through a kernel density estimate, generated by the \texttt{smoothScatter} function in the \texttt{R} programming language. The density gradient ranges from white (low density) to red (high density), and the 500 largest points are plotted in black.  

\subsubsection{Are samples with large influence sketch scores particularly likely to be mislabeled?}\label{results2}

\begin{figure*}
\centering
\includegraphics[height=5in]{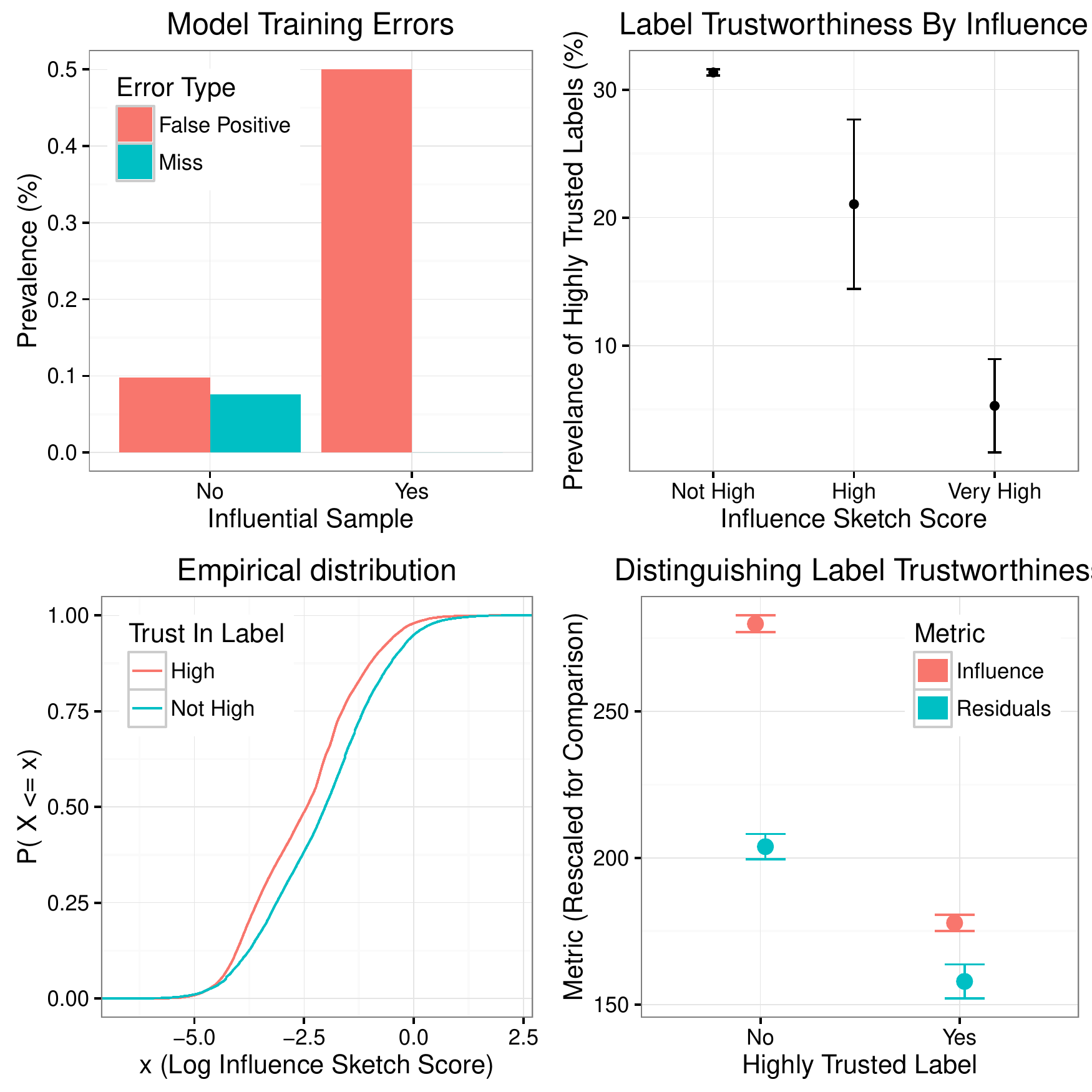}
\caption{ \emph{Samples with large influence sketch scores are particularly likely to be mislabeled.}   The errors bars reflect $\pm 2$ standard errors for estimating means. See Section~\ref {results2} for more detail.}
\label{megaplot}
\end{figure*}

In Figure~\ref{megaplot} (top left), we show the distribution of nominal model training errors\footnote{The discussion in this section relies on the nomenclature of Section~\ref{errors}. } for influential vs. non-influential samples.   We note first that nominal training accuracy is lower for influential samples (99.41\%) than non-influential samples (99.81\%), suggesting that the model struggles more when fitting the influential samples.  Moreover, whereas the model is well-calibrated overall (for non-influential samples, the percentage of nominal false positives, $0.098\%$, and nominal misses, $0.076\%$, is nearly equal), the model shifts dramatically towards committing false positives on the influential samples (for influential samples, the percentage of nominal false positives is $0.58\%$, whereas the model actually makes \emph{no} nominal misses.)    This finding is extremely interesting with regards to detecting mislabeled files.   Our belief about signature-based detection, which is commonplace (see, e.g. ~\cite{kantchelian}), is that if a file is indicted, it is likely actually malicious; however, many malicious files do not get caught by signature-based detection, especially initially, because it takes time to identify suspect files and to write signatures for them.  In other words, the labeling mechanism can be assumed to have a high miss and a low false positive rate.   This would in turn cause the modeling process to have a high nominal false positive and low nominal miss rate, precisely the pattern observed with the influential samples.    This result suggests that influential miscategorized samples may be especially likely to be mislabeled.  

To further pursue the relationship between influence and mislabeling, as discussed in Section~\ref{Data}, label trustworthiness judgments were obtained (as a proxy for mislabeling probabilities) for a subset of nominally good samples in the training set.  In Figure~\ref{megaplot} (bottom left), we see that samples whose nominal good labels are not highly trusted have higher influence scores.  In Figure~\ref{megaplot} (top right), we see that the percentage of nominally good samples with highly trusted labels drops, in a statistically significant way, for sets of samples with high levels of influence.   The 95\% confidence intervals for the true prevalence of highly trusted labels is $31.34\% \pm 2.3\% $    for the non-influential group, $21.05\% \pm 6.6\% $ for the high influence group, and $5.29\% \pm 3.6\% $ for the very high influence group.  In Figure~\ref{megaplot} (bottom right), we show that influence scores are more discriminating of label mistrust than residual scores (which are more straightforward to calculate).   The blue points represent the mean Pearson squared residuals and the red points represent influence sketch scores, where both metrics have been rescaled for the purposes of visualization.  The plot reveals that influence scores are more discriminating than residual scores -- the signal to noise ratio is substantially higher for the influence sketch scores than for the residuals.   

To more directly investigate whether influence sketching can help point towards mislabeled samples, we manually analyze the 10 most influential nominal mistakes  by the model (here, all nominal false positives) to determine an actual ground truth for these samples.    Our manual analysis revealed that, despite the nominal labels of good, 8/10 of these samples indeed had ``dark grey" to ``black" properties.  A number of these samples appeared to represent newly discovered malware variants (undetected by the labeling mechanism).  For example, the particular version of the video editing tool infected with Parasitic Ramnit passed scans by ESET, Microsoft, Sophos, and Trend.

\section{Why find influential samples} \label{why_do_it}
Our analysis reveals some uses for the identification of influential samples in large-scale regressions. In particular, we have focused on manually analyzing influential samples which were nominally miscategorized by the model. 

\begin{enumerate}
\item The influential \emph{nominally} miscategorized, but \emph{actually} correctly categorized, samples can be an excellent place to find mislabeled samples.  As manual analysis is expensive, focus should be placed on critical samples with the largest influence on the model's fitted weights and downstream predictions. Thus, compared to traditional label cleansing approaches which recommend investigating all miscategorized samples, the influential miscategorizations provide the largest ``bang for your buck."    In this case, by investigating the nominal false positives from the top 1\% most influential files, we would reduce our manual labors from investigating 1,780 false positives to only 133 false positives -- an over 13-fold reduction in labor efforts -- and we would know that continuing further would yield diminishing returns on our efforts.   Moreover, the results in Section~\ref{results2} suggest that nominal miscategorizations, when influential, are in fact particularly likely to be mislabeled. 

\item The influential \emph{nominally} miscategorized, and \emph{actually} miscategorized, samples can point to key model vulnerabilities and spur ideas for improving the regression model, e.g. through novel feature generation.  Two samples -- the Pharos Popup client and CryptDrive -- were labeled by the regression model as bad, despite ``good" training labels and the fact that these samples were actually legitimate software.  
Do these samples perhaps have ``confusing" combinations of features? Might it be helpful to add new features to help more generally clarify the status of printer drivers or installers for data encryption software?   We note that, whereas in our particular dataset, influential samples were all nominal false positives, in the general case, a dataset would yield influential nominal misses.   Whenever a manual analysis determines that these influential nominal misses are actual misses, then influence sketching would point towards key model vulnerabilities that could be exploited by an adversarial attacker.  

\end{enumerate}





\section{Conclusion}

We show that the influence sketching algorithm, which embeds random projections into the construction of a generalized Cook's distance score, can successfully flag samples that are especially impactful on the performance of large-scale regressions.    This finding suggests that influence sketching can be useful for constructing priority queues for manual analysis by experts. Moreover, further statistical analysis and the case study suggest that influential apparent model miscategorizations can point towards mislabeled samples, which, in the cybersecurity domain, can lead to undiscovered malware.   Finally, in other contexts (where influential misses are prevalent), influence sketching may highlight critical model vulnerabilities open to exploitation by an adversary.   Future research should further explore this connection between influence sketching and adversarial attacks, as well as the potential for generalizing influence sketching to more general models (e.g., neural networks).  

\appendix

As a case study, we manually analyze the 10 most influential nominal model miscategorizations (all nominal false positives).   Despite the ``good" training labels, 8/10 samples had ``dark grey" to ``black" properties.  These samples were:
\begin{enumerate}
\item A Russian-language GUI for sending fraudulent SMS messages from any international phone number.   The messages can contain links which download malware, go to fake login pages, or hijack the recipient's phone.   Interestingly, this program's code checks for the existence of Avast Anti-Virus.  
\item DarkMailer v1.38i, a very fast bulk emailer which taps into a network of zombie computers to send up to 500,000 emails per hour. 
\item A video editing tool which was infected with Parasitic Ramnit.   Because of the Ramnit worm, each time this file is run, it can infect more files: Windows executable files, HTML files, Microsoft Office files, etc.
\item A Chinese-language chat program that is abused by a malware parasitic. 
\item LanAgent, an employee monitoring software which tracks websites visited and email correspondences to ``detect activities that have nothing to do with work and will show you how efficiently your employees spend their office hours."  
\item A version of MySQL Manager 2008 cracked by "=iNViSiBLE=-", a known piracy group.  This software acts as if it is validly licensed without the user needing to have a license.   This kind of modification creates security vulnerabilities which make it very easy to add a backdoor. 
\item XRUMER, a search engine optimization program which illicitly registers and post to forums with the goal of boosting search engine rankings.  This program can bypass security techniques for preventing spam, such as  email activation before posting, CAPTCHAs, and security questions (e.g. what is 2+2?).   The program uses HTTP proxies in order to anonymize the source IP to make it more difficult for administrators to block posts. 
 \item  traceapi.dll, a shared library originally developed for Microsoft Detours Express,  which is used to hook functionality of a process in order to supply back different functionality.  This is a \emph{Potentially Unwanted Program}; it was developed for a clean use case, but can be used to abuse another process. 
\end{enumerate}

Additionally, 2/10 samples were actually ``good". These could be considered influential model mistakes, pointing to potential deficiencies in feature repesentation: 
 \begin{enumerate}
\item An installer for CryptDrive, which can perform data encryption.
\item Pharos Popup client, a printer driver for the public cloud often used in academic settings. It is legitimate software, but contains API calls that are, in conjunction, sometimes associated with malicious behavior.
\end{enumerate}


\end{document}